\documentclass{article}

\usepackage[preprint, nonatbib]{neurips_2022}

\usepackage[utf8]{inputenc} 
\usepackage[T1]{fontenc}    
\usepackage{hyperref}       
\usepackage{url}            
\usepackage{booktabs}       
\usepackage{amsfonts}       
\usepackage{nicefrac}       
\usepackage{microtype}      
\usepackage{xcolor}         

\usepackage{amsmath}
\usepackage{amssymb}
\usepackage{array}
\DeclareMathOperator*{\argmax}{argmax}

\newcommand{\LL}{\mathcal{L}}
\newcommand{\M}{\mathcal{M}} 
\newcommand{\E}{\mathcal{E}}
\newcommand{\D}{\mathcal{D}}
\newcommand{\G}{\mathcal{G}}

\newcommand{\Z}{\mathcal{Z}}

\usepackage[pdftex]{graphicx}
\usepackage{tikz}

\title{Intrinsic Motivation in Model-based Reinforcement Learning: A Brief Review}

\author{%
  Artem~Latyshev \\
  Moscow Institute of Physics and Technology\\
  Moscow, Russia \\
  \texttt{latyshev.ak@phystech.edu} \\
  \And
  Aleksandr I.~Panov \\
  Moscow Institute of Physics and Technology \\
  Federal Research Center “Computer Science and Control” of Russian Academy of Sciences \\
  AIRI \\
  Moscow, Russia \\
  \texttt{panov@airi.net} \\
}

\begin{document}

\maketitle

\begin{abstract}
  The reinforcement learning research area contains a wide range of methods for solving the problems of intelligent agent control. Despite the progress that has been made, the task of creating a highly autonomous agent is still a significant challenge. One potential solution to this problem is intrinsic motivation, a concept derived from developmental psychology. This review considers the existing methods for determining intrinsic motivation based on the world model obtained by the agent. We propose a systematic approach to current research in this field, which consists of three categories of methods, distinguished by the way they utilize a world model in the agent's components: complementary intrinsic reward, exploration policy, and intrinsically motivated goals. The proposed unified framework describes the architecture of agents using a world model and intrinsic motivation to improve learning. The potential for developing new techniques in this area of research is also examined.
\end{abstract}

\section{Introduction}

Reinforcement learning (RL) methods demonstrate impressive results in many problems of generating artificial agent behavior. They show efficient processing at the human level in games \cite{mnih_human-level_2015}, in classic tasks with a large search space \cite{silver_mastering_2017}, in the control of robotic systems \cite{schulman_proximal_2017}. At the same time, such algorithms already outperform experts on specific tasks, but how they learn from a scalar reward environment signal is far from how a person learns. It is well established that human beings not only pay attention to the outcome of a task, but also rely on their understanding of cause-and-effect relationships in the surrounding environment and their previously acquired skills. With these universal capabilities and knowledge, a person can quickly learn to solve complex problems and reuse the accumulated knowledge.

On the one hand, in RL, there are many methods using world model learning \cite{moerland_model-based_2022}, which allows the agent to store and generalize knowledge about the dynamical properties of the environment. On the other hand, even with a model, the agent forms a policy based on a task-specific reward signal so that in its absence, no learning occurs. One of the solutions is suggested by analogies with human learning and the psychology of motivation: \emph{human behavior determined by intrinsic motivation \cite{ryan_intrinsic_2000} leads to effective learning in the absence of extrinsic drives (such as reward signals for RL agent)}. The adoption of intrinsic motivation has led to a new approach in the development of artificial agents \cite{oudeyer_what_2007, baldassarre_intrinsically_2013, aubret_survey_2019, aubret_information-theoretic_2022}, where the key element is the intrinsic reward signal, which serves as a substitute for rewards in RL.

The intrinsic motivation approach suggests a method for obtaining information about the environment through the manipulation of the agent. Thus, it provides task-agnostic learning of the world model. Having access to a model allows for the identification of a wide variety of intrinsic motivation signals. According to them, the exploration policy is trained, the internal goals are formed, and their priority is determined. The world model and intrinsic motivation complement each other.

The objectives of the article are to review existing model-based RL algorithms that use methods of intrinsic motivation; to propose a unified framework that systematizes the accumulated knowledge in this field; to consider ways for further development.

The structure of the review is as follows. Sections \ref{sec:background} and \ref{sec:agent_learning} describe the formal statement of the problem. In Section \ref{sec:interaction_model_im}, we present our primary contribution by examining techniques for determining model-based intrinsic motivation methods. We group these methods into three research directions: complementary intrinsic reward (Section \ref{subsec:additional_reward}), exploration policy (Section \ref{subsec:expl_policy}), and intrinsically motivated goals (Section \ref{subsec:intr_goals}). For each of them, specific examples are analyzed and the impact on the learning process is presented. In Section \ref{sec:problems} we consider the problems facing the improvement of the current architectures of intrinsically motivated agents and directions of future development.

\section{Background}
\label{sec:background}

The Markov Decision Process (MDP) is formally defined by the set $\langle S, A, T, \rho, \gamma \rangle$. For each state in the environment $s \in S$ and possible action $a \in A$ the transition function $T$ and the reward function $R$ are given. The transition function $T: S \times A \rightarrow p(S)$ determines the conditional probability $p(s_{t+1}|a_t, s_t)$ of getting into the next state $s_{t+1} \in S$ from the current one $s_t \in S$ after execution action $a_t \in A$ in the environment. The reward function $R: S \times A \times S \rightarrow \mathbb{R}$ evaluates a scalar characteristic (reward signal) of each transition triple $\tau = (s_t, a_t, s_{t+1})$. The initial state of the agent in the environment $s_0 \in S$ is from the corresponding distribution $\rho: S \rightarrow p(S)$. The factor $\gamma \in [0, 1]$ is a discounting parameter.

The policy $\pi$ is responsible for the agent's behavior --- the function $\pi: S \rightarrow p(A)$ that defines the probability $p(a_t|s_t)$ of the action execution in the current state $s_t$.

Reinforcement learning methods solve the optimization problem of maximizing expected return:

\begin{equation}
    \pi^* = \argmax_\pi \mathbb{E}_{\substack{
        s_0 \sim \rho(s_0),
        a_t \sim \pi(a_t|s_t), \\
        s_{t+1} \sim p(s_{t+1}|s_t, a_t)}} \left[\sum_{t=0}^\infty \gamma^t R(s_t, a_t, s_{t+1}) \right].
    \label{eq:mdp_opt}
\end{equation}

\subsection{Model-based RL}

There are two main approaches to the problem of sequential decision-making: reinforcement learning and planning \cite{moerland_model-based_2022}. The model-based RL (MBRL) combines them. \emph{MBRL considers a class of MDP algorithms that use a world model (planning part) and store a global solution (RL part) \cite{moerland_model-based_2022}}. Here the global solution is the optimal policy $\pi^*$ (eq. \ref{eq:mdp_opt}). The world model $\M$ is a set of functions that describes the dynamics in the environment $\E$.

The model can be obtained by learning or known in advance. In the first case, learning is based on the collected experience of interaction with the environment. In the second case, no training is required. The last option is very rare in real problems since in most cases the environment dynamics are unknown. In the following discussion, the world model is assumed to be obtained by learning unless otherwise stated.

\begin{figure}[h]
    \centering
    \includegraphics[width=\linewidth]{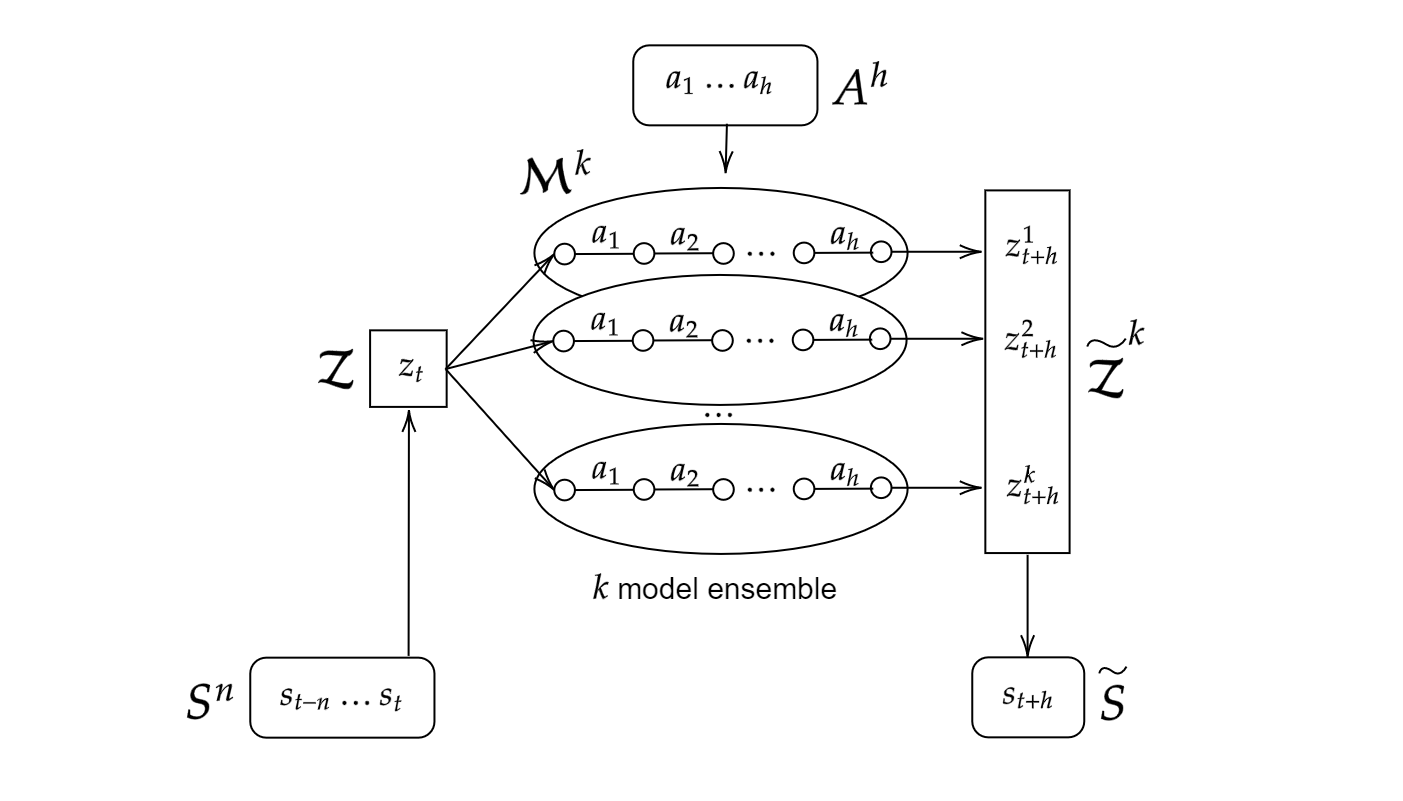}
    \caption{The structure of the model. $S^n$ --- the history of $n$ states. $\Z$ --- the latent space, a special case is then $\Z$ matches $S$. $\M^k$ --- an ensemble of $k$ individual models. $A^h$ --- a sequence of $h$ actions. $\Z^k$ --- the result of the model execution. $\Tilde{S}$ --- the state after $h$ steps from the initial one.}
    \label{fig:fullmodel}
\end{figure}

The formal description of the environment dynamics has a lot of variations. However, in all of them, there is a transition from one state to another when performing actions in the environment. The possible components of the model determining the specific one are presented in Fig. \ref{fig:fullmodel}.

The model makes predictions of the next agent state $\Tilde{S}$ based on the history of $n+1$ previous states $S^n$\footnote{$n>0$ means partial observability of the environment, there the history is necessary for determining Markov states.} and the sequence of $h$ actions $A^h$. To construct such a transition a special latent space $\Z$ is considered. It makes it possible to distinguish essential features for the structure of the environment and allows performing transitions in a simpler space than the original state space. The transitions discussed above correspond to the forward dynamics model. However, the inverse dynamics model is also contained in the scheme, but only $A^h$ is calculated from $S^n$ and $\Tilde{S}$.

The standard way is to use one model $\M$ \cite{pathak_curiosity-driven_2017}, but also the ensemble $\M^k$ of several models is widespread \cite{hafez_deep_2019, mendonca_discovering_2021, pathak_self-supervised_2019, shyam_model-based_2019}. In the latter case, each separate part acts as an independent model that makes its own prediction. These predictions can be used as is or combined into the final result.

\subsection{Intrinsic motivation}

In RL many interesting and innovative ideas have been borrowed from the psychological theories of human behavior. One such example is the concept of intrinsic motivation. According to the founders of the theory of self-determination \cite[p. 3]{ryan_intrinsic_2000}: \emph{"Intrinsic motivation is defined as the doing of an activity for its inherent satisfactions rather than for some separable consequence"}.

An intrinsically motivated person enjoys the activity without extrinsic stimuli, which is the opposite of extrinsic motivation. This one is completely determined by teacher praise, monetary rewards, and fear of punishment. In reinforcement learning, the main idea of which is based on the signal of reward and punishment, extrinsic motivation is the foundation of learning. However, it was shown in \cite{barto_intrinsic_2005} that intrinsic motivation is also naturally contained in the mathematical formalism of computational reinforcement learning. In this case, a special reward signal is introduced --- intrinsic reward $R_{int}$.

The calculation of $R_{int}$ can use not only the transition triple $\langle s_t, a_t, s_{t+1} \rangle$, as it happens for external rewards, but also on internal representations $\Z$ and model $\M$. A detailed discussion of variations in the definition of the intrinsic reward signal is presented in Section \ref{subsec:intr_signal}.

\section{Intrinsically motivated agent learning}
\label{sec:agent_learning}

\begin{figure}[h]
    \centering
    \includegraphics[width=\linewidth]{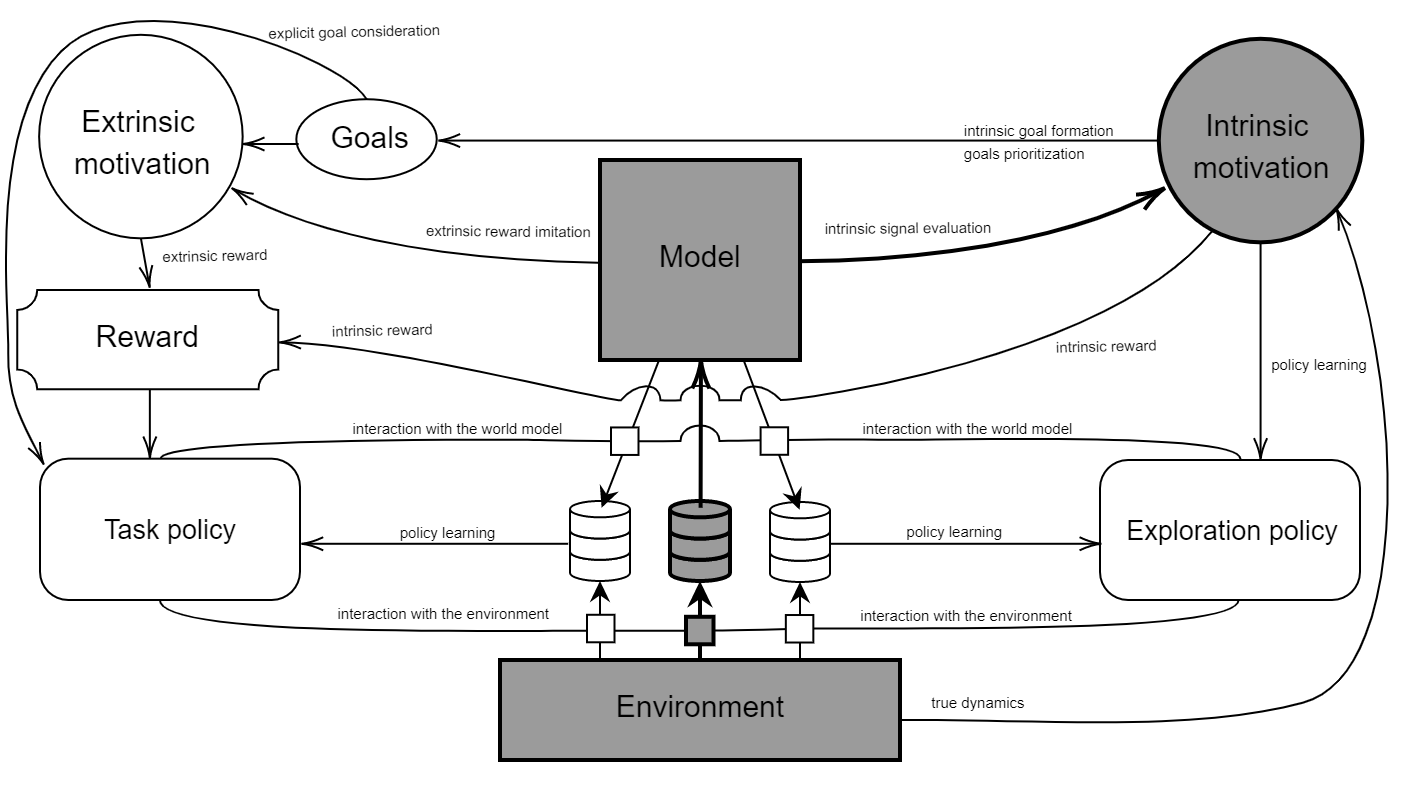}
    \caption{The scheme of interacting components when an agent is trained by intrinsic motivation methods based on the world model. Highlighted elements --- the basis of the approach under consideration.}
    \label{fig:fullagent}
\end{figure}

In intrinsically motivated reinforcement learning both extrinsic and intrinsic motivation signals are considered. It becomes essential to determine the way these two approaches interact with each other and with other parts of the agent - model, and policy (see Fig. \ref{fig:fullagent}).

From the left side of the diagram (see Fig. \ref{fig:fullagent}) there is an extrinsic motivation determined by the task goal. It is the main controller of task policy learning for the given MDP. From the right side, intrinsic motivation is determined by the characteristics and interaction of the world model and the environment. This type of motivation provides the agent with:
\begin{itemize}
    \item a complementary intrinsic reward $R_{int}^*$ that corrects the main task reward from the environment, which affects the learning of the target strategy;
    \item an exploration policy $\pi_\epsilon$, guided by which the agent collects experience in the environment for learning the entire system;
    \item a set of intrinsic goals $\G_{int}$ and a schedule for learning them, that is task-agnostic and significantly increases the degree of agent autonomy.
\end{itemize}

Not every existing method of intrinsic motivation implements all three components $R_{int}^*, \pi_{\epsilon}, \G_{int}$ at once. Table \ref{table:intr_methods} (for description of the “Signal type” column see Section \ref{subsec:intr_signal}) contains information about the presence or absence of each of them in the agents considered in the review.

An agent learning includes learning of its model and two policies, which is determined as the optimization problem (see Section \ref{subsec:losses} for details). The main source of information in this case is the agent's interaction with the world. By choosing certain actions in the current state, the agent as a result receives training examples, which are usually stored in memory $\D$ --- the source of the training set.

\begin{table}
  \caption{The presence of components used by methods of intrinsic motivation based on the world model. $\chi[\M]^*$ --- the scalar motivational signal is not calculated, but the structure of the world model is used.}
  \label{table:intr_methods}
  \centering
  \begin{tabular}{l|cccc}
    \toprule
    Method name & $R^*_{int}$ & $\pi_\varepsilon$ & $G_{int}$ & Signal type\\
    \midrule
     SelMo\cite{groth_is_2021} & --- & --- & + & $\LL[\M]$ \\
     ICM\cite{pathak_curiosity-driven_2017} & + & --- & --- & $\LL[\M]$ \\
     EMI\cite{kim_emi_2019} & + & --- & --- & $\LL[\M]$ \\
     Plan2Explore\cite{sekar_planning_2020} & --- & + & --- & $\LL[\M]$ \\
     LEXA\cite{mendonca_discovering_2021} & --- & + & + & $\LL[\M]$ \\
     MEEE\cite{yao_sample_2021} & + & --- & --- & $\LL[\M]$ \\
     MAX\cite{shyam_model-based_2019} & --- & + & --- & $\LL[\M]$ \\
     AWML\cite{kim_active_2020} & --- & + & --- & $\Delta\M$ \\
     VIME\cite{houthooft_vime_2017} & + & --- & --- & $\Delta\M$ \\
     Deep ICAC\cite{hafez_deep_2019} & + & --- & --- & $\Delta\M$ \\
     GDE\cite{volpi_goal-directed_2020} & + & --- & --- & $\chi[\M]$ \\
     WRW\cite{mezghani_walk_2022} & --- & + & + & $\chi[\M]$ \\
     EC\cite{savinov_episodic_2019} & + & --- & --- & $\chi[\M]$ \\
     CEE-US\cite{sancaktar_curious_2022} & --- & + & --- & $\LL[\M]$ \\
     Director\cite{hafner_deep_2022} & --- & --- & + & $\LL[\M]$\\
     CC-RIG\cite{nair_contextual_2020} & --- & --- & + & $\chi[\M]^*$ \\
     SMORL\cite{zadaianchuk_smorl_2020} & --- & --- & + & $\chi[\M]^*$\\
     SRICS\cite{zadaianchuk_self-supervised_2022} & --- & --- & + & $\chi[\M]^*$ \\
    \bottomrule    
  \end{tabular}
\end{table}

\subsection{Training set}
\label{subsec:buffers}

\begin{figure}[h]
    \centering
    \includegraphics[width=\linewidth]{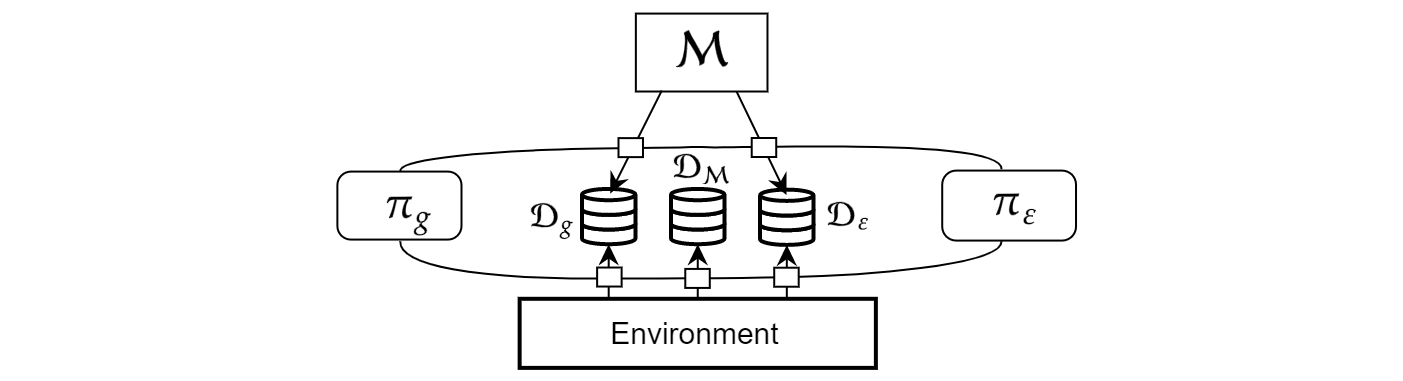}
    \caption{Data collection for training. Task and exploration policies collect data for learning from the world model and environment into the memory $\D_g, \D_\epsilon, \D_\M$.}
    \label{fig:buffers}
\end{figure}

An agent has three learning components: task policy $\pi_g$, exploration policy $\pi_\epsilon$, and model $\M$. In general, for each of them, there is its memory (training set): $\D_g$, $\D_\epsilon$, $\D_\M$, respectively (see Fig. \ref{fig:buffers}). The memory stores trajectories of the agent $\tau_H = (s_i, a_i, s_{i+1})_{0:H-1}$ consisting of $H$ transitions from one state $s_i$ to another $s_{i+1}$ under the condition of action $a_i$ execution. In this case, the action is sampled from the policy, and the next state is sampled from the transition dynamics.

To fill the memory, a policy interacts with the environment (to store examples of transitions from the true dynamics), the model ("dreaming," when the learned model determines the transitions), or mixed (some trajectories are determined by the model and the other by the environment):

\begin{equation}
\begin{gathered}
    \D_g = \{\tau_H = (s_i, a_i, s_{i+1})_{0:H-1}| a_i \sim \pi_g(s_i), s_{i+1} \sim [\M, T](s_i, a_i)\},\\
    \D_\epsilon = \{\tau_H = (s_i, a_i, s_{i+1})_{0:H-1}| a_i \sim \pi_\epsilon(s_i), s_{i+1} \sim [\M, T](s_i, a_i)\}.
\end{gathered}
\end{equation}

To learn the model of the world, only trajectories generated by the true dynamics of the environment can be used since the model's purpose is to learn exactly it. However, the agent can perform actions by the exploration policy and by task one or even by mixing the trajectories of both:

\begin{equation}
    \D_\M = \{\tau_H = (s_i, a_i, s_{i+1})_{0:H-1}| a_i \sim [\pi_g,\pi_\epsilon](s_i), s_{i+1} \sim T(s_i, a_i)\}.
    \label{eq:model_buffer}
\end{equation}
\noindent

But it is worth noting that the main objective of the task policy is to solve the MDP for the initial task. The exploration policy, at the same time, helps to obtain the most significant amount of information from the world. It has a greater variety of transitions in trajectories than task one, which improves model learning.

\subsection{Loss functions}
\label{subsec:losses}

In the formal statement of the learning problem, an optimization problem is defined to find the minimum of a certain function often expressed as the loss function $\LL$. Thus, each of the main agent components has its own functional $\LL_g, \LL_\epsilon, \LL_\M$. However, it is possible to simultaneously optimize the entire function of the system that combines corresponding components. Arguments of the loss functions for the exploration policy, the task policy, and the world model are displayed on the agent diagram (see Fig. \ref{fig:fullagent}). Let's consider each of the cases in detail.

\paragraph{The task policy} naturally requires a pair of states and actions for learning and the reward signal. The reward can be determined entirely by extrinsic or complementary signals defined by intrinsic motivation. Such mixing of different motivations into one is a standard method of combining. However, it introduces a bias in the MDP problem defined for the task, based only on extrinsic rewards. The reward implicitly transfers information about the task to the policy, but the latter can explicitly depend on the goal. Thus, the loss function for the task policy can be represented:
\begin{equation}
    \LL_g = \mathbb{E}_{\tau_H \sim \D_g} l(\tau_H, [R,R_{int}], g, \pi_g),
\end{equation}
\noindent
where $\D_g$ --- the memory (see Section \ref{subsec:buffers}), $g$ --- the goal, $[R, R_{int}]$ is the combined reward signal, and $l$ --- one of the standard reinforcement learning loss functions such as discounted return (see eq.~(\ref{eq:mdp_opt})).

\paragraph{The exploration policy} solves a similar optimization problem, but it relies solely on intrinsic rewards, unlike task one. Such a policy is task-agnostic since intrinsic motivation does not seek to achieve the goal. Then the loss function is:
\begin{equation}
    \LL_\epsilon = \mathbb{E}_{\tau_H \sim \D_\epsilon} l(\tau_H, R_{int}, \pi_\epsilon),
    \label{eq:loss_pi_e}
\end{equation}
\noindent
where $l$ is one of the standard functions for reinforcement learning objectives, such as task policy.

\paragraph{The world model} learning procedure is different from learning policies since it solves the problem of supervised learning rather than reinforcement. In the former, the true examples are already known. Policies in the environment collected them. It is only needed to learn how to reproduce them. The variety of the loss functions for the world model is excellent, but the loss function mainly calculates the difference between the predictions of the model and true examples from the training set:
\begin{equation}
\begin{gathered}
    \LL_\M = \mathbb{E}_{\tau_H \sim \D_\M} l_M(\tau_H, \M),\\
    \LL_\M = \mathbb{E}_{\tau_H \sim \D_\M} l_F(\M(S^n, A^h),\Tilde{S}),\\
    \LL_\M = \mathbb{E}_{\tau_H \sim \D_\M} l_I(\M(S^n,\tilde{S}), A^h),
\end{gathered}
\end{equation}
\noindent
where $S^n, A^h, \Tilde{S}$ are the state, actions, and predicted states from the $\D_\M$, $l_M$ is a function that determines the difference between prediction and truth (e.g., the sum of the squared difference for vector states), $l_I$ and $l_F$ are the loss functions of the inverse and forward dynamics model respectively.

\section{Intrinsic motivation and model}
\label{sec:interaction_model_im}

Learned world model contains the agent's knowledge about the dynamics of the surrounding world. On the one hand, this makes it possible to reduce the number of interactions with the environment for learning the policy. The data from the model compensates for the decrease in the information flow from the environment. On the other hand, direct access to the transition dynamics through the model makes it possible to simplify the exploration problem.

\begin{figure}[h]
    \centering
    \includegraphics[width=\linewidth]{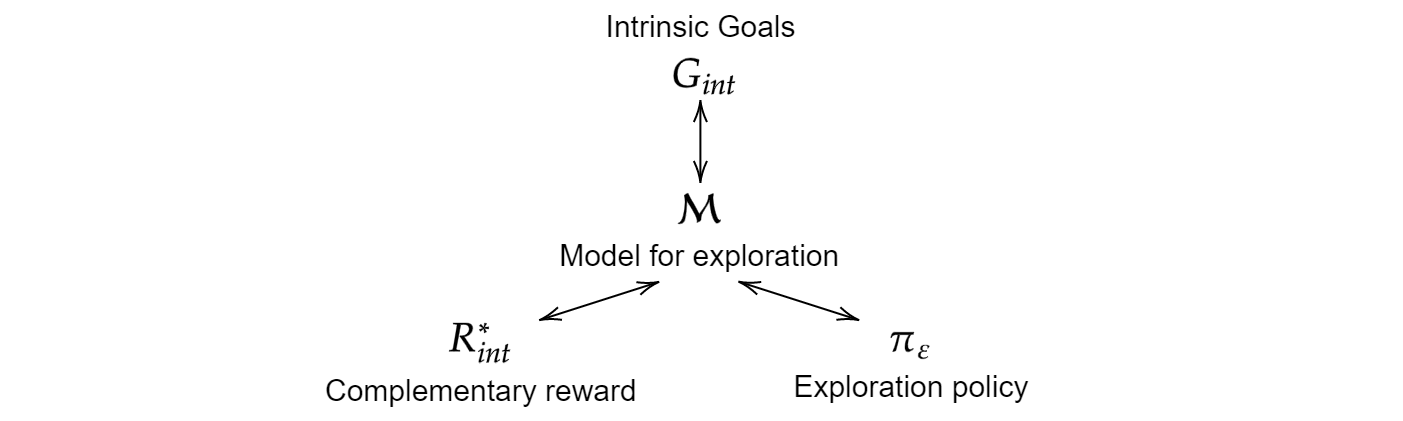}
    \caption{Model and intrinsic motivation.}
    \label{fig:model_explore}
\end{figure}

The application of a model for exploration\footnote{Here exploration is considered in a broad sense: there is a search not only for possible states in the environment but also for skills --- behavioral policies for various goals.} is an area of direct contact with methods of intrinsic motivation, that consists of three approaches distinguished by the influence on the task policy (see Fig. \ref{fig:model_explore}):

\begin{itemize}
    \item the first is based on modifying the reward $R$ for the agent (the intrinsic reward is added to task one) (see Section \ref{subsec:additional_reward});
    \item the second is based on changing the data acquisition into memory $\D$ by using an exploration policy (see Section \ref{subsec:expl_policy});
    \item the third is based on setting goals $g$ (determined by signals of intrinsic motivation, as well as by the structure of the world) as learning problems for the task policy $\pi_g$ (see Section \ref{subsec:intr_goals}).
\end{itemize}

In each of these three ways, there is an intrinsic motivation signal. Its maximization provides the agent with exploratory capabilities.

\subsection{Intrinsic signal}
\label{subsec:intr_signal}

A feedback signal is needed to train the policy. It numerically is characterized by the quadruple $\langle s_t, a_t, s_{t+1}, g \rangle$. There are two main approaches to evaluate this signal in the intrinsic motivation methods: \emph{knowledge-based}, which takes into account only states and actions $\langle s_t, a_t, s_{t+1} \rangle$, and \emph{competence-based}, which characterizes both intrinsic goals and the possibility of achieving them $\langle s_t, a_t, s_{t+1}, g \rangle$. Such a division of methods does not reflect the specifics of using the model to determine intrinsic motivation.

One of the agent's objectives is to search for information about the real dynamics of the environment to train its world model. To perform such exploration, markers are needed to indicate the success of the process. Such markers --- intrinsic motivation signals --- use
\begin{itemize}
    \item the current uncertainty of the model, i.e., some model error;
    \item the knowledge gained of the model from the received data;
    \item the morphology of the environment, which makes it possible to identify important transitions, actions, or states.
\end{itemize}

\paragraph{Model uncertainty $\LL[\M]$} is the most comprehensive group of methods to determine the intrinsic motivation signal:
\begin{equation}
\begin{aligned}
    \LL[\M]: R_{int} &=
        \begin{cases}
        |\Tilde{S}-\M(S)|, & \text{between the model and true dynamics};  \\
        D[\M^k(S)], & \text{between ensemble components,}
        \end{cases}
\end{aligned}
\label{eq:lm}
\end{equation}
\noindent
where $|\ldots-\ldots|$ denotes the difference between elements, and $D[\dots]$ defines the diversity of elements (e.g., deviation, variance). One of the intrinsic reward signals is the error of the predicted next state by comparing it with the true one (see eq. (\ref{eq:lm}); ICM \cite{pathak_curiosity-driven_2017}, SelMo \cite{groth_is_2021}, EMI \cite{kim_emi_2019}, Director \cite{hafner_deep_2022}). However, this approach requires constant access to the real state. Another way is to determine the variance of ensemble predictions --- see eq. (\ref{eq:lm}) (Plan2Explore \cite{sekar_planning_2020}, LEXA \cite{mendonca_discovering_2021}, MEEE \cite{yao_sample_2021}; MAX \cite{shyam_model-based_2019} --- the signal is similar but uses the Jensen-Shannon divergence).

\paragraph{Knowledge gain $\Delta\M$} --- signals that determine the change in the model when new information is received:
\begin{equation}
    \Delta\M: R_{int}(t) = |\M(t) - \M(t-n)|,
\end{equation}
\noindent
where $\M(t)$ is the model state at time $t$, and $n$ is the horizon of time steps over which we track the change in the model. Examples of such signals are the difference of loss function in several steps (AWML $\gamma$-Progress \cite{kim_active_2020}), Kullback-Leibler divergence between current and updated model predictions (VIME \cite{houthooft_vime_2017}), ensemble prediction improvement (Deep ICAC \cite{hafez_deep_2019}).

\paragraph{Environment morphology $\chi[\M]$} defines a set of reward signals that characterizes the structural properties of the world:
\begin{equation}
    \chi[\M]: R_{int} = X[\M, S],
\end{equation}
\noindent
where $X$ is a specific function that makes it possible to characterize one of the morphological properties of the environment numerically, such a signal is not explicitly associated with the process of learning of either the policy or the world model since there is no comparison: the models with each other, the models at different time steps. This intrinsic motivation signal characterizes the morphology of the environment to mark states or actions that are important for the particular environment. For example, the empowerment \cite{klyubin_all_2005, volpi_goal-directed_2020} defines the reward as the information capacity between the sequence of actions $A^h$ and the subsequent state $\Tilde{S}$. Another example is the reachability signal, which helps to form intrinsic goals for the agent \cite{mezghani_walk_2022, savinov_episodic_2019}.

\subsection{Intrinsic reward as a complement to extrinsic}
\label{subsec:additional_reward}

Determining the reward signal is one possible way to influence the performance of the task policy. Extrinsic reward is the main component that determines the target behavior of the agent. However, researchers have many problems with defining such a signal that prevents the agent from stagnation in local optima and provides learning in a reasonable time. Closely related to this is the sparse reward problem, when the agent cannot progress in learning because it does not receive a feedback signal. One way to overcome this difficulty is to complement a rare extrinsic reward with an auxiliary dense signal --- intrinsic reward (e.g., solving the problem of passing the game "Montezuma's Revenge" \cite{burda_exploration_2018}).

The mixing of extrinsic and intrinsic signals is a linear combination of rewards:
\begin{equation}
    [R,R_{int}]: r = R + \alpha R_{int},
    \label{eq:lin_intr_extr}
\end{equation}
which is used in the classical version of policy learning (see eq. (\ref{eq:loss_pi_e})). For example, such a method is implemented in \cite{pathak_curiosity-driven_2017, houthooft_vime_2017, kim_emi_2019}. However, it is possible to combine value functions --- intrinsic and extrinsic --- instead of raw rewards (e.g., the MEEE algorithm \cite{yao_sample_2021}), which saves collected information about the extrinsic task in the value function separately from the exploratory signal.

The mixed reward signal has a bias for the given MDP. To eliminate this effect, it is necessary to select an adaptive coefficient $\alpha$ (see eq. (\ref{eq:lin_intr_extr})), which decays over time, or the intrinsic signal should converge to zero as the agent learns. The latter property is true for $\LL[\M]$\footnote{In some cases, the prediction error will give a constant reinforcing signal to uncontrolled noise in the environment, e.g., the noisy TV problem \cite{pathak_curiosity-driven_2017}.} and $\Delta\M$ signals.

\begin{figure}[h]
    \centering
    \includegraphics[width=0.75\linewidth]{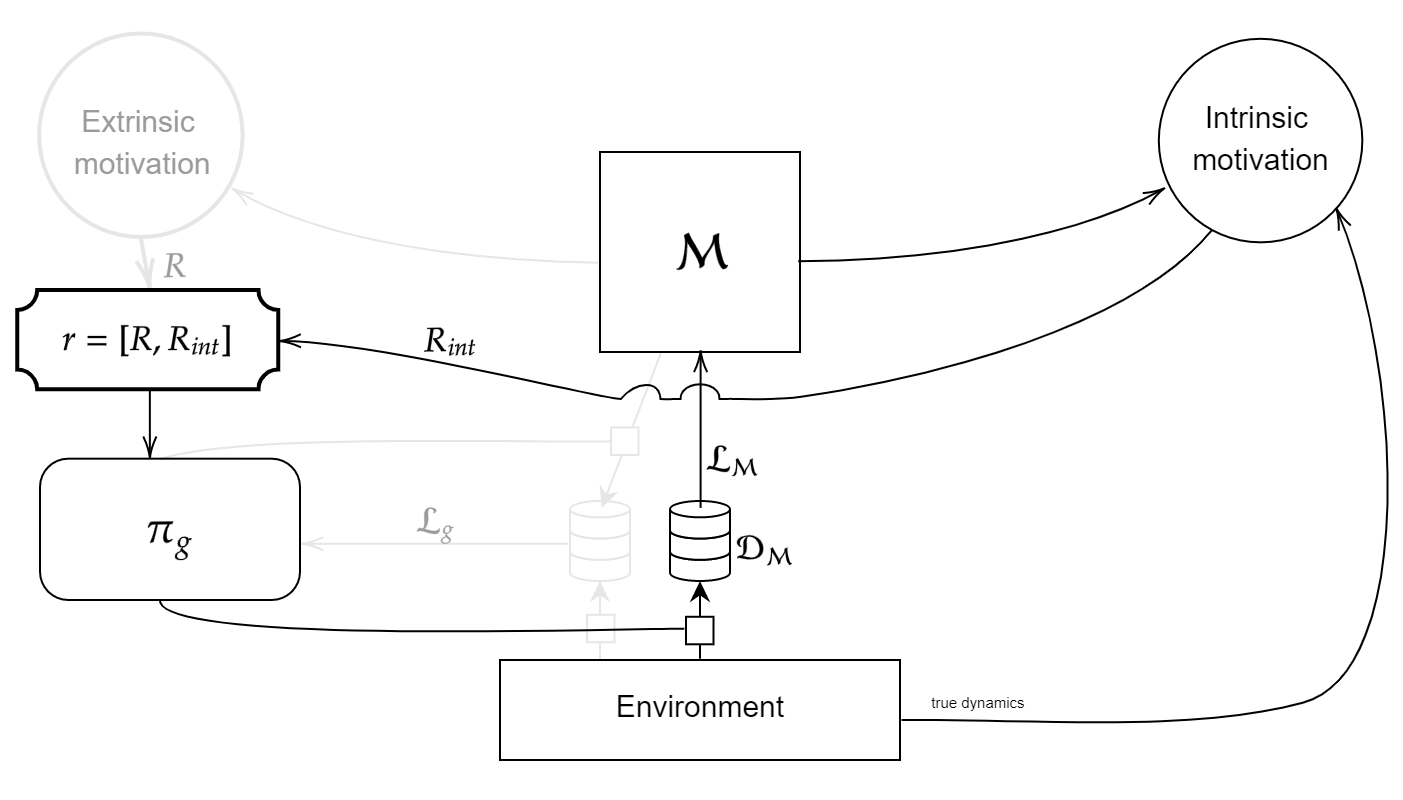}
    \caption{The scheme of the world model application through the reward to the training of the agent.}
    \label{fig:model_reward}
\end{figure}

The scheme of model application through complementary reward to improve exploration consists of several components (see Fig. \ref{fig:model_reward}). The world model generates an intrinsic reward. The intrinsic motivation signal, mixed with task one, trains the task policy and provides it with an exploratory component. This component helps the agent to get the necessary data in the $\D_\M$ memory (see eq. (\ref{eq:model_buffer})). The model is used until it becomes sufficiently well trained (that the intrinsic motivation signal drops to zero) or until the task policy becomes successful (heuristic for the adaptive coefficient $\alpha$).

\subsection{Exploration policy}
\label{subsec:expl_policy}

Mixing an intrinsic signal with an extrinsic reward binds the learning of the policy with the learning of the world model, which can become a problem. The model trained in such a way is biased in the context of achieving a specific goal. Therefore, it needs additional fine-tuning on new tasks. The indicated problem is resolved by the insertion of an additional exploration policy that is not related to task one. The only objective of which is to generate training data that does not depend on the specific goal given to the agent.

The problem of learning an exploratory policy (see Section \ref{subsec:losses}) is defined similarly to learning a task policy, except that the reward is determined only by intrinsic motivation.

The exploration policy can work alone, training the agent model later used for specific tasks in the environment. Thus, an agent with a world model can learn to achieve the goal set by extrinsic motivation without interacting with the environment and immediately performs it at an acceptable level. For example, in \cite{sekar_planning_2020} and \cite{shyam_model-based_2019}, exploration policies (Plan2Explore and MAX respectively) collection of data in $\D_\M$ is based on the variance of the ensemble of models. Moreover, the training of the policy itself occurs according to the model --- the memory $\D_\epsilon$ is determined by the model $\M$. The CEE-US agent \cite{sancaktar_curious_2022} works according to a similar scheme, but it uses an ensemble of graph networks. The task policy can also train the model --- collecting the $\D_\M$ memory. Usually, the proportion of samples from different policies is an agent hyperparameter (see \cite{mendonca_discovering_2021}).

In addition to training the model to solve previously unknown tasks, the exploration policy can be used as an initialization for the task policy. For example, in \cite{groth_is_2021}, the authors propose to save copies of the exploration policy during learning to use them as a set of pretrained skills.

\begin{figure}[h]
    \centering
    \includegraphics[width=0.75\linewidth]{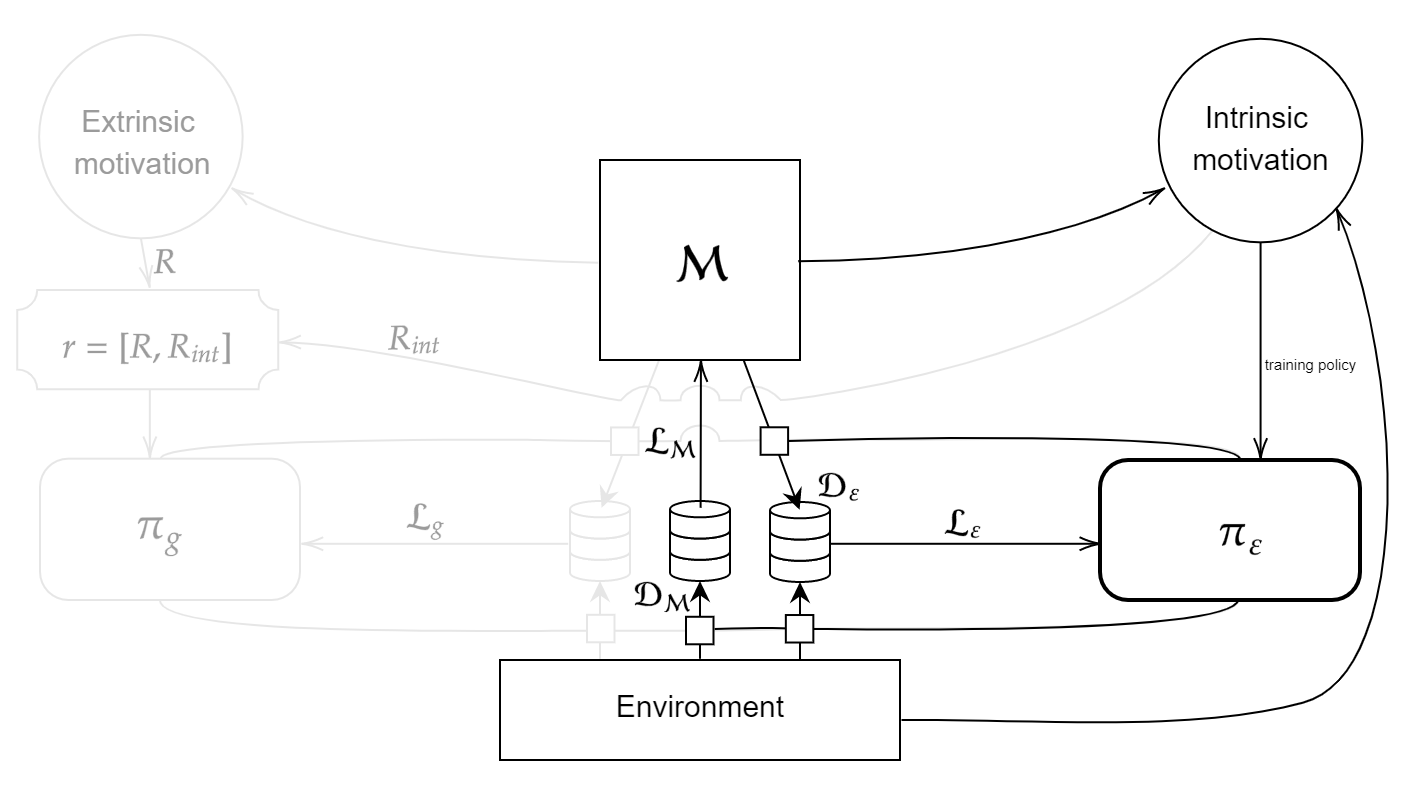}
    \caption{The scheme of the model application through the exploration policy to the training of the agent.}
    \label{fig:model_policy}
\end{figure}

Thus, the model of the world is involved in the agent training through the exploration policy as follows below (see Fig. \ref{fig:model_policy}). The model trains the exploration policy on intrinsic reward and data generated by the model. This policy collects the data in the most relevant environment for training the world model. A side application of the exploration policy can be the initialization of the task policy and the definition of intrinsically motivated goals (e.g., the LEXA agent \cite{mendonca_discovering_2021}).

\subsection{Intrinsically motivated goals}
\label{subsec:intr_goals}

To train an agent to achieve goals, it is necessary to define the space of goals and set a schedule for their selection. The standard way to create a goal space is to make it the same as a state space. The choice of goals can be represented as some MDP: the agent has a hierarchical structure, where at the lower level, the policy $\pi(a|s,g)$ selects actions, and at the next level, the policy $\pi(g|s)$ selects the goals that determine the behavior of the lower level. This representation makes it possible to transfer the standard methods of learning policy with a model to train the upper-level policy by intrinsic rewards (Director \cite{hafner_deep_2022}).

Many agents do not use such a homogeneous hierarchical representation: they define goals from the history of the agent's interaction with the environment or world model. In this case, the policy learning algorithm consists of two subtasks. The first is to collect a set of goals or construct a space of goals. The second is to sample targets from the previously formed set.

\paragraph{Formation of goals set.} The model of the world is capable of defining a set of goals that improve it. Here goals are states in which the model makes poor predictions. For example, an agent uses an exploratory policy trained on the data from a model to acquire a set of states that become potential targets for the policy (LEXA \cite{mendonca_discovering_2021}).

The world model can also select the entire set of states that the agent can achieve, on the basis of information about their reachability, so as not to try to learn the impossible. This approach usually uses a simplified model version, which determines the probability of reaching one state from another in some fixed number of steps (e.g., \cite{mezghani_walk_2022}). Another way is to bind the representation of the goals with the states from which they are achievable (e.g., CC-RIG \cite{nair_contextual_2020}).

An algorithmically convenient representation of transitions and states in the environment can be formed using the model. It highlights the morphology of the world, from which a set of goals is determined without unnecessary challenges. For example, you can use an object-oriented representation of states when their vector is formed based on raw sensory information, the components of which correspond to the characteristics of individual objects (as implemented in the SMORL algorithm \cite{zadaianchuk_smorl_2020}). Or the model can define an interaction graph with separate objects as nodes. This method makes it possible to decompose the achievement of a common goal into subgoals. Each of them is an intrinsic goal for the agent (e.g., the SRICS algorithm \cite{zadaianchuk_self-supervised_2022}).

\paragraph{The choice of goals} is primarily determined by the same principles that formed the space of goals. When forming a set of goals, a numerical selection criterion is used (e.g., the probability of reaching, see above), which naturally corresponds to the priority of choosing a goal\footnote{It is worth noting that this is not the only possible signal, but one that uses information from the model of the world. There are other signals, for example, the success of the agent in performing the goal, see review \cite{oudeyer_what_2007}.}. Or the constructed interaction graph specifies the learning schedule for individual nodes (SRICS \cite{zadaianchuk_self-supervised_2022}).

Thus, the model of the world makes it possible through intrinsic reward (see Fig. 7 arrow A) or through an exploration policy (see Fig. 7 arrow B) to choose goals for further learning. Also, the model can help to choose goals that are important in a particular environment based on the trained world structure (see Fig. 7 arrow C).

\begin{figure}[h]
    \centering
    \includegraphics[width=0.75\linewidth]{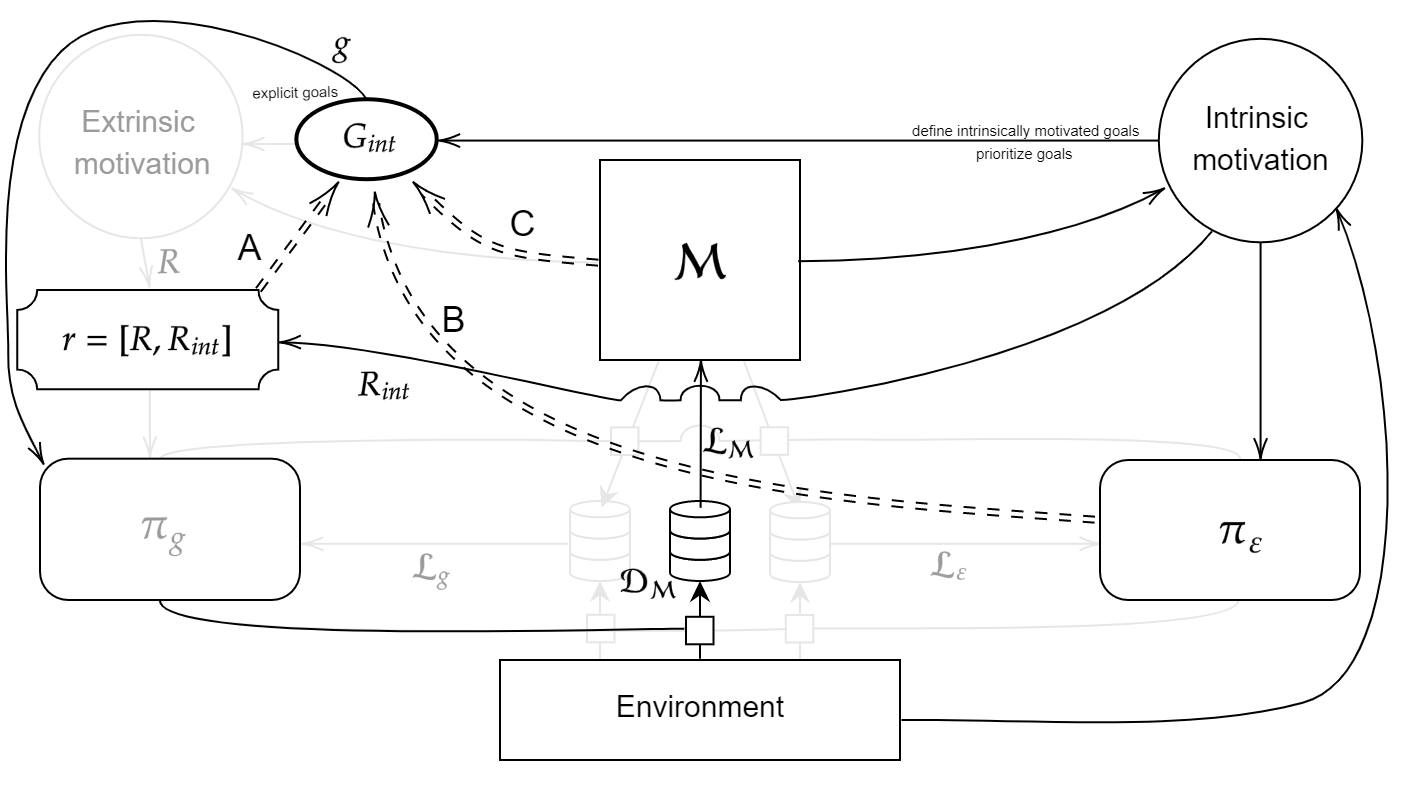}
    \caption{The scheme of the model application through setting goals to the agent's training. Dashed arrows reveal the relationship between goal setting and model-based intrinsic motivation.}
    \label{fig:model_goal}
\end{figure}

\section{Challenges and problems}
\label{sec:problems}

The main application of the intrinsic motivation methods is to solve some problems in reinforcement learning related to exploring the environment and the agent capabilities (see review \cite{aubret_survey_2019}). These problems are sparse reward problems, construction of the latent space, abstract skills formation, and training schedule.

Reinforcement learning algorithms perform well when the agent receives a dense reward, i.e., for almost every completed action. However, in some cases, such a signal is absent or \emph{sparse} (e.g., "Montezuma's Revenge" \cite{bellemare13arcade}). Intrinsic motivation methods suggest using a specific exploratory policy (see Section \ref{subsec:expl_policy}) or adding a dense intrinsic reward signal that directs the agent to perform exploration. Moreover, the trained world model corrects the behavior, considering the already-known knowledge about the environment.

In deep reinforcement learning \emph{latent representations} created by artificial neural networks according to the reward signal are task-specific, which causes difficulties in applying the trained agent model in tasks with another reward signal. Intrinsic motivation provides independent learning aimed at obtaining useful information from the environment and building representations that take into account the environment dynamics but not the specifics of the task (e.g., similar methods were considered when discussing the formation of a goal space, see the review \cite{aubret_survey_2019} for more details).

Training an agent with skills\footnote{some abstract actions that certain group behaviors to achieve an intermediate goal} defines two more problems: \emph{the formation of skills} and \emph{learning schedule}. Among the works on intrinsic motivation, this area of research is referred to as competence-based motivation. Research studies \cite{oudeyer_what_2007, aubret_survey_2019} consider this type of motivation in detail and \cite{forestier_intrinsically_2022} proposes the IMGEP approach taken it as a basis. As discussed above, one of the ways to form goals, which in turn explicitly defines the corresponding skill, is to use a model of the world. So, the relationships between the representations formed in the model make it possible to determine the sequence of learning goals.

Intrinsic motivation solves many RL problems. However, there are problems and limitations with intrinsic motivation methods themselves. A lot of intrinsic reward signals are based on the prediction error. If the environment is stochastic, such an error may be irremovable, which means it will attract the agent. However, it does not improve his behavior, for example, a noisy TV \cite{burda_exploration_2018}. Using an exploratory policy raises the question of when and how long exploration motives should determine the agent's behavior. Many methods of intrinsic motivation, especially those that do not use the model of the world, provide exploration only in the short term. Still, the model-based approach with the planning solves this drawback.

Others can solve problems faced by some methods of intrinsic motivation. So, for example, to choose between exploration and exploitation, each of the behaviors can be represented as a separate skill and trained in this paradigm (see Section \ref{subsec:intr_goals}). Thus, the integration of intrinsic motivation approaches into one whole system is one of the promising tasks that have not yet been solved in science. However, there are some works in this direction (IMGEP \cite{forestier_intrinsically_2022}, Vygotskian Artificial Intelligence \cite{colas_vygotskian_2022}).

\section{Conclusion}
\label{sec:conclusion}

This article reviews the existing methods of intrinsic motivation that form a model of the world for learning. On the one hand, the presence of a model increases the learning performance in the environment on account of additional experience. However, this approach is standard practice for many reinforcement learning algorithms unrelated to intrinsic motivation. On the other hand, the model, as a source of experience accumulated by the agent, determines the intrinsic motivation that guides the agent in exploration.

The framework is presented that systematizes the considered methods. Three ways of applying the world model determine the main classes that make up the proposed classification. The first is based on \emph{complementing the main reward} with intrinsic model signals. Among them, we have identified three types: uncertainty $\LL[\M]$ (usually model prediction error), knowledge gain $\Delta\M$ (change in the model itself when new information is received), and environment morphology $\chi[\M]$ (internal relationships between elements of the model of the world). The second class of methods relies on the application of the model to train \emph{the exploration policy}. And the third class contains methods that form \emph{the goals and the sequence of their achievement}, based on the signals of the model and its structural properties. 

Intrinsic motivation algorithms often implement only one of the ways to apply the model (see Table~\ref{table:intr_methods}) without using all its potential. The consistent coordination of complementary intrinsic rewards, exploration policy, and intrinsic goals sets the direction for our future research.

\bibliographystyle{plain}
\bibliography{bibliography}

\end{document}